\documentclass[sigconf, nonacm]{acmart} 
\usepackage{bbm}
\usepackage{enumitem}

\AtBeginDocument{%
  \providecommand\BibTeX{{%
    \normalfont B\kern-0.5em{\scshape i\kern-0.25em b}\kern-0.8em\TeX}}}

\begin{document}

\title{On the Effect of Robot Errors on Human Teaching Dynamics}


\author{Jindan Huang}
\affiliation{%
  \institution{Tufts University}
  \city{Medford}
  \state{Massachusetts}
  \country{USA}}
\email{jindan.huang@tufts.edu}

\author{Isaac Sheidlower}
\affiliation{%
  \institution{Tufts University}
  \city{Medford}
  \state{Massachusetts}
  \country{USA}}
\email{isaac.sheidlower@tufts.edu}

\author{Reuben M. Aronson}
\affiliation{%
  \institution{Tufts University}
  \city{Medford}
  \state{Massachusetts}
  \country{USA}}
\email{reuben.aronson@tufts.edu}

\author{Elaine Schaertl Short}
\affiliation{%
  \institution{Tufts University}
  \city{Medford}
  \state{Massachusetts}
  \country{USA}
}
\email{elaine.short@tufts.edu}

\begin{abstract}

Human-in-the-loop learning is gaining popularity, particularly in the field of robotics, because it leverages human knowledge about real-world tasks to facilitate agent learning. When people instruct robots, they naturally adapt their teaching behavior in response to changes in robot performance. While current research predominantly focuses on integrating human teaching dynamics from an algorithmic perspective, understanding these dynamics from a human-centered standpoint is an under-explored, yet fundamental problem. Addressing this issue will enhance both robot learning and user experience. Therefore, this paper explores one potential factor contributing to the dynamic nature of human teaching: robot errors. We conducted a user study to investigate how the presence and severity of robot errors affect three dimensions of human teaching dynamics: feedback granularity, feedback richness, and teaching time, in both forced-choice and open-ended teaching contexts. The results show that people tend to spend more time teaching robots with errors, provide more detailed feedback over specific segments of a robot's trajectory, and that robot error can influence a teacher's choice of feedback modality. Our findings offer valuable insights for designing effective interfaces for interactive learning and optimizing algorithms to better understand human intentions.
\end{abstract}



\maketitle

\section{Introduction}

As robots become increasingly integrated into daily life, they will need to learn from and engage with non-expert users. To enable this, human-in-the-loop learning allows users to leverage their intuitive understanding of real-world tasks, facilitating robot learning more effectively than pure trial-and-error methods. While human teachers will adapt to the teaching interface and the robot learner, not all teaching methods and robot behaviors are straightforward for the teachers to learn and respond to.  Therefore, it is crucial to design algorithms and systems that leverage the natural approaches people use for teaching to both enhance the teaching experience and appropriately react to a teacher's feedback. 

\begin{figure}[t]
    \centering
    \includegraphics[width=0.5\textwidth]{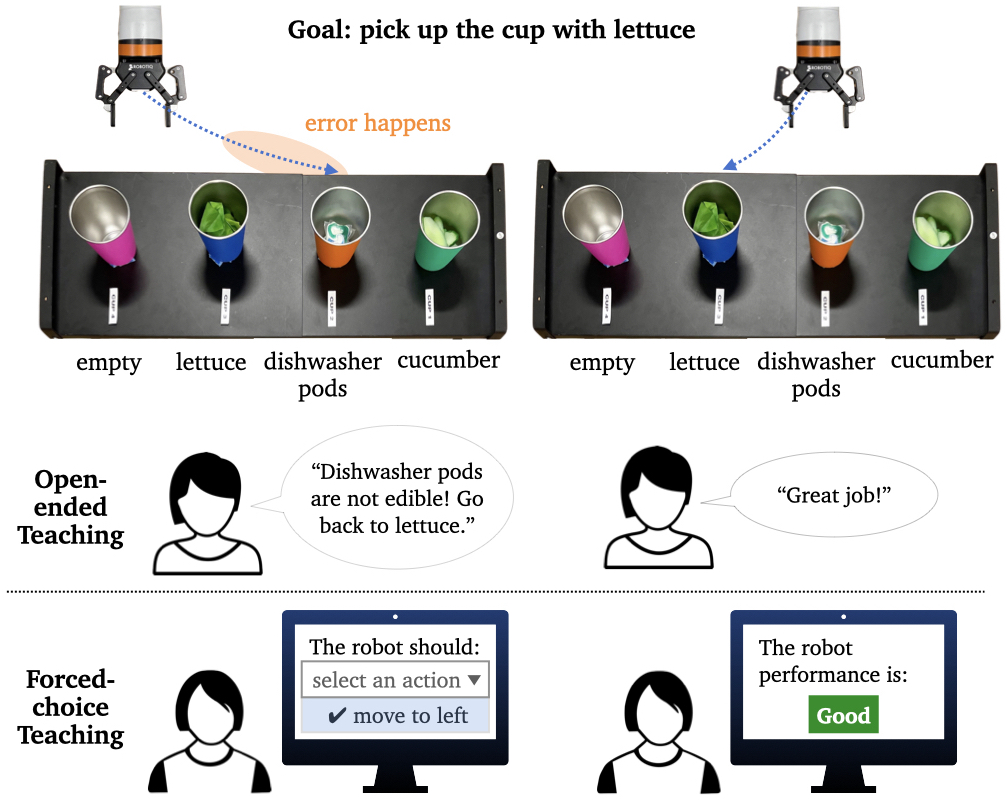}
        \caption{In both forced-choice and open-ended teaching contexts, people adapt their teaching behavior based on the presence or absence of robot errors.}
    \label{fig:front_img}
\end{figure}

When people teach, they communicate different intentions through a variety of teaching signals \cite{loftin2014strategy} and accommodate their teaching based on the performance of the learner \cite{thomaz2008teachable}. Much of existing work has considered the dynamic nature of human teaching from an algorithmic perspective. In particular, work has focused on how to collect and process various sources and forms of human feedback for agent learning, such as demonstrations or evaluations \cite{biyik2022learning}. These algorithmic approaches often assume that only the content of the feedback changes. However, understanding this dynamic from a fundamentally human-centered perspective (in terms of how users decide and convey their teaching choices) still remains in its early stages. While some research has identified common behavioral variations in human teaching \cite{huang2024modeling}, the factors contributing to these variations and their mechanisms are under-explored. 

In this paper, we focus on one potential factor: robot errors. Human teachers will inevitably encounter robot errors as they instruct a robot in a new task. Previous studies have shown that robot errors can influence general human behavior, such as social responses \cite{kontogiorgos2020behavioural}, engagement with robots \cite{van2019take} and perceptions of robots \cite{tian2021taxonomy}. There is a need to study the relationship between robot errors and dynamic human teaching as it can have significant implications as to how to design both robot learning algorithms and human teaching interfaces. For example, depending on the interface, the same feedback may be elicited from robot errors that greatly differ in severity. This may not lead to efficient robot learning, as more severe errors may require the robot to learn more information. On the other hand, if the interface is too complicated or too open-ended, the user may find it cumbersome to respond to various errors differently. Or, if there is a strong relationship between error severity and the modality with which the user tends to give feedback, we can leverage that relationship to design more efficient learning algorithms and better teaching experiences. Therefore, understanding human teaching dynamics in response to robot errors is an important step.

Our work shows that teaching dynamics \emph{are} influenced by robot errors and that those influences can be predictable. We validated this claim through an empirical study focused on the online, Reinforcement Learning from Human Feedback (RLHF), teaching setting. Specifically, we investigated how the presence and severity of robot errors affect three dimensions of human teaching dynamics: feedback granularity, feedback richness and teaching time, in both forced-choice and open-ended teaching contexts. Our findings reveal that people tend to spend more time teaching the robot with errors, and give more detailed feedback over a more specific segment of a robot trajectory. Furthermore, in forced-choice teaching, when a user has the option between instructing the robot through an action correction or evaluating the robot's behavior through binary feedback, they would significantly more often choose to instruct when the robot makes an error, and evaluate when the robot is successful. These results offer valuable implications for designing expressive interactive teaching systems and developing learning algorithms that can leverage nuances in human teaching.

\vspace{-4pt}

\section{Background}

Human-in-the-loop (HITL) learning is an approach that integrates human input and inherent expertise to facilitate agent learning. It has garnered significant attention in research due to its potential to improve agent performance across various real-world domains \cite{mosqueira2023human, retzlaff2024human}. For example, reinforcement learning from human feedback (RLHF) is a HITL learning paradigm that has proven effective for training or refining models, especially with online crowdsourced feedback \cite{christiano2017deep, kaufmann2023survey}. In this paper, we focus on an RLHF setting while taking into account findings from previous HITL learning research. 

Traditional HITL learning approaches usually restrict feedback to a single type \cite{xiao2020fresh, dai2023safe} or treat variation in human feedback as noises (e.g., using Boltzmann distributions \cite{jeon2020reward}). However, recent work suggests that they may oversimplify the intricacies of human teaching and overlook valuable information given by human instructors \cite{lindner2022humans, cui2021understanding, casper2023open}, which potentially leads to deterioration in agent learning performance. In an attempt to address these challenges, researchers have begun taking into account the ways in which people naturally approach the task of teaching, such as adapting teaching based on the learner's performance by using multiple teaching signals and strategies. However, current research focuses primarily on the technical challenges of incorporating diverse feedback sources into agent learning \cite{cruz2018multi, ghosal2023effect} and developing standardized encoding format for various feedback types \cite{metz2023rlhf, yuan2024uni}. There is a growing need to understand a fundamental question: \textit{what factors influence dynamic human teaching behavior towards robots?}

One factor likely to influence teaching behavior is errors from learners. In human-human teaching, when a student makes some mistake, people often switch among multiple teaching strategies such as clarification, correction and motivation \cite{du2001modelling, oppenheim2021mental}. This shift occurs because errors prompt teachers to estimate the severity \cite{hyland2006teachers} and reassess their perception of the student's abilities \cite{fisher1986twenty}. Similarly, in human-robot teaching, human instructors exhibit different teaching styles when interacting with robots with varying performance \cite{kaochar2011towards} and adjust their teaching behavior accordingly as they dynamically update their impressions of robot capabilities \cite{thomaz2006reinforcement}. 

Robot errors could also result in this dynamic human-robot teaching. For instance, Chi et al. \cite{chi2023calibrated} found that as social agents made fewer inappropriate responses, human teachers provided less frequent instruction. However, most work related to robot errors focuses on general human-robot interaction contexts. Previous research has demonstrated that the presence of robot errors can impact user engagement \cite{kim2009people, aliasghari2021different} and people's perceptions on robots such as reliability and trust \cite{nesset2021transparency, ye2019human, washburn2020robot}. Moreover, the severity of robot mistakes affects both human-robot collaboration performance \cite{salem2015would, garza2018failure, stiber2024forging} and human reaction intensity \cite{brooks2016analysis, mirnig2017err}. As HITL robot learning becomes more prevalent, understanding how robot errors shape human teaching is increasingly critical. Therefore, this paper investigates the role of robot errors as a key factor influencing human-robot teaching dynamics.

\section{Methodology} 

Feedback dynamics vary in many ways. Users might chose a different modality, e.g. evaluation, instruction, or demonstration; or they might adjust the specificity of their feedback, such as critiquing a single robot action or an entire trajectory. These variations may be a function of many things such as fatigue, preference, or the setting of the learning task. In this work, we focus on how robot error affects teaching dynamics in an online, RLHF-style teaching scenario. To do this, we design a within-subjects study and apply it to two different contexts: open-ended teaching and forced-choice teaching. Through these studies, we examine how robot errors can impact three important dimensions of a human teacher's feedback dynamics: \emph{feedback granularity}, \emph{feedback richness}, and \emph{teaching time}.

\emph{Feedback granularity} refers to how specific a user's feedback is with respect to time. For example, a user may wish to critique a whole robot trajectory or a specific moment where the robot did something particularly good or bad. Feedback granularity as defined in this way is often predefined by the designers of the system as the either a window in the past for which the feedback is assigned to (this is known as ``credit assignment'' \cite{sutton1984temporal}), or implicitly determined by showing end-users trajectories of fixed length to elicit their preferences \cite{park2022surf}. A user, however, may want to able to be specific about what part of the robot's behavior their feedback is referring to. Thus, to study this dynamic from a teacher's perspective, we explicitly design an interface which a user can specify a credit assignment window. 

\emph{Feedback richness} refers to how much information is present in a piece of feedback. Intuitively, this represents the difference between broad judgements about a behavior, such as saying "good/bad job" and specific instructions about how to do better, such as "next time move a bit to the right." The concept of different levels of feedback richness often shows up in both robot training scenarios \cite{wilde2020improving} and recommendation systems \cite{jannach2021survey}. For example, after providing binary feedback to an agent, the teaching interface may query the user for more specific information. While more rich feedback may help the robot learn better, always querying for rich feedback may cause excessive fatigue in the teacher \cite{lin2020review}. Thus, understanding when users prefer to give more or less rich feedback, in this case as a function of robot error, can provide insights into how to improve both the teaching experience and learning performance. 

\emph{Teaching time} refers to how long a user spends teaching a robot in response to a given trajectory. For any given robot behavior, there may be one or more aspects of that behavior the user wishes to provide feedback for. For example, if a robot makes multiple mistakes in a single attempt at a task, a user may provide separate feedback for each individual error or take the time to formulate a single piece of broader feedback over that whole trajectory. Likewise, if the robot is doing particularly well, a user may take the time to provide the robot with extra praise. On the other hand, if the user has been teaching the robot for a while and is relatively satisfied with the robot's behavior they may choose to provide more concise feedback. Each of these circumstances can be captured by measuring teaching time. Thus, teaching time can give insights into both how easy it was for the user to understand and formulate their thoughts about the robots behavior as well as how bad or good the robot behavior may have been.

Based on these metrics, we developed three hypotheses for both forced choice and open-ended teaching contexts. The hypotheses are as follows: 
\begin{itemize}[nolistsep]
\item \textbf{H1: Influence on feedback granularity}
\begin{enumerate}[]
    \item Individuals tend to offer feedback with lower granularity for erroneous trajectories compared to successful ones;
    \item The severity of a robot error is correlated with the granularity of feedback provided by individuals;
\end{enumerate}
\item \textbf{H2: Influence on feedback richness} 
\begin{enumerate}[]
    \item Individuals tend to provide more rich feedback for erroneous trajectories compared to successful ones;
    \item The severity of a robot error is correlated with the richness of feedback provided by individuals;
\end{enumerate}
\item \textbf{H3: Influence on teaching time} 
\begin{enumerate}[]
    \item Individuals allocate more time to providing feedback for erroneous trajectories compared to successful ones;
    \item The severity of a robot error is correlated with the amount of time people spent on teaching.
\end{enumerate}
\end{itemize}

\subsection{Experiment Setup} 
\subsubsection{Environment}
\begin{figure}[b]
    \centering
    \includegraphics[width=0.4\textwidth]{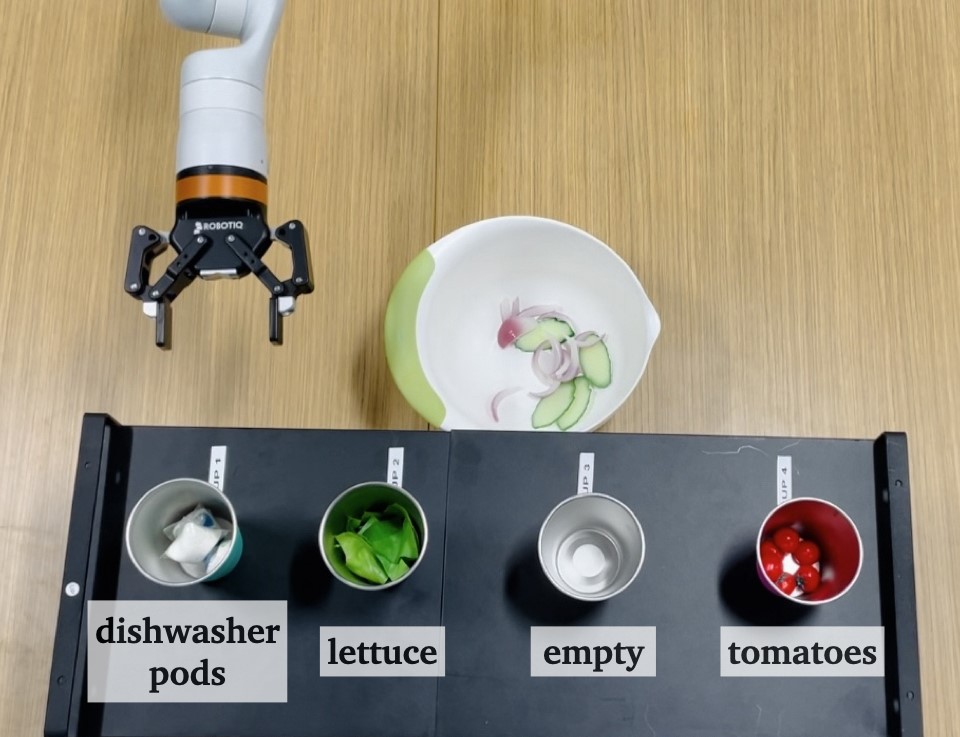}
        \caption{Salad preparation study environment. The robot arm is tasked with picking up the correct salad ingredient and pouring that ingredient into the bowl.}
    \label{fig:study_env}
\end{figure}

We implemented a robot environment using Kinova Gen3 arm for salad preparation (see Fig. \ref{fig:study_env}). The goal of the robot arm is to pick up the cups with necessary ingredients based on the given recipe and pour out the ingredients from those cups into a mixing bowl in the center of the table. There are 4 cups in the environment, and each of them can contain useful ingredients (e.g., lettuce, cucumber, tomatoes, onion), inedible dishwasher pods, or is empty.

\subsubsection{Robot Trajectories \& Error Severity.}
\label{sec:robot_errors}
To understand how robot errors influence human teaching, we designed 6 robot trajectories and recorded them as videos: 4 with errors and 2 without. The length of all videos is $20\pm2 $ seconds. Two successful videos are: 
\begin{itemize}
    \item \textit{add lettuce}: Add lettuce to the mixing bowl correctly
    \item \textit{add tomatoes}: Add tomatoes to the mixing bowl correctly
\end{itemize}

Each erroneous trajectory includes one of the following mistakes: 
\begin{itemize}
    \item \textit{add dishpods}: Add dishwasher pods to the mixing bowl 
    \item \textit{add nothing}: Add nothing to the mixing bowl as a result of grabbing and attempting to pour from an empty cup
    \item  \textit{miss tomatoes}: Miss grabbing the cup with tomatoes by failing to close the gripper 
    \item  \textit{drop onion}: Drop onion pieces outside of the mixing bowl as a result of attempting to pour out the onions only partially over the mixing bowl
\end{itemize}

These errors are informed by a prior user study \cite{aliasghari2021effect}, which offers a comprehensive list of robot mistakes with corresponding severity rankings within the context of food preparation. We ensured that a range of error severity was covered across the four failure trajectories, from a relatively low severity error,  such as failing to add an ingredient by forgetting to close the robot gripper, to a relatively high severity error,  such as adding cleaning product to the salad.

In our study, participants assessed the severity of robot errors by rating the perceived robot performance in each trajectory using a 5-point Likert scale (1-excellent, 5-very poor). Despite our prior assumptions over how severe each error may be, by directly querying participants for their perception of the error severity, we can account for any unexpected individual differences in that perception. Furthermore, this rating is collected after participants provide their feedback to ensure that it is not perceived as part of the robot teaching process.

\subsubsection{Interactive teaching interface.} 
\label{sec:teaching_interface}
\begin{figure*}[t]
    \centering
    \includegraphics[width=0.75\textwidth]{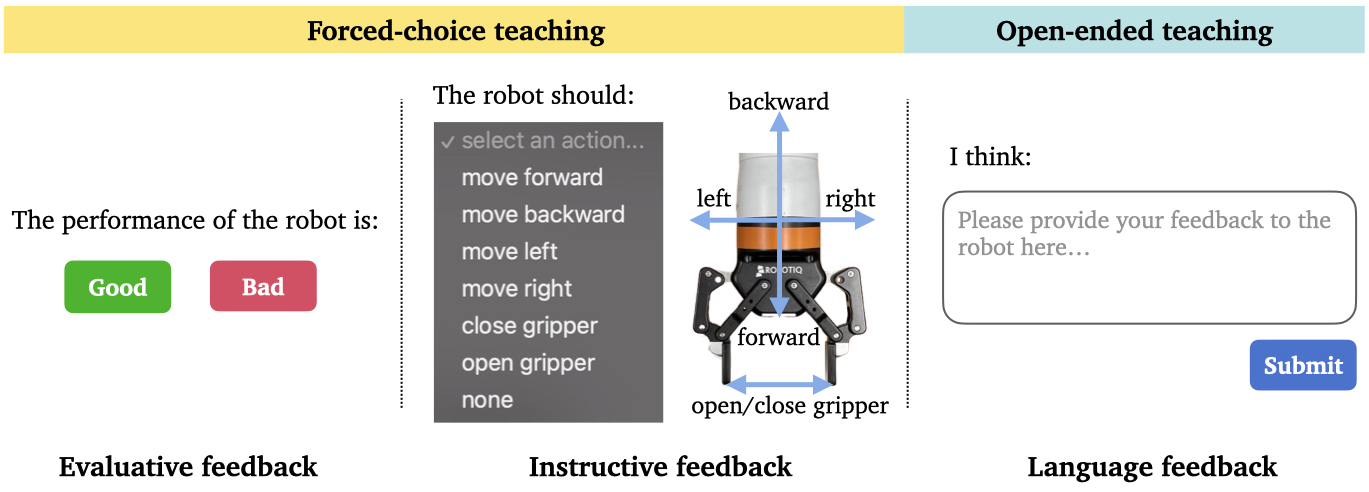}
        \caption{Types of feedback used in our forced-choice teaching and open-ended teaching contexts}
        \label{fig:teaching_feedback}
\end{figure*}

We developed a user interface that prompts participants to provide feedback for improving robot performance (see Fig. \ref{fig:teaching_interface}). Since different query styles may affect people's perception on robot intelligence \cite{cakmak2012designing}, we used a universal prompt to collect feedback from human teachers (\textit{"Please help the robot perform better"}). Participants were asked to guide the robot using either \textit{structured evaluative/instructive feedback} or \textit{language feedback} (see Fig. \ref{fig:teaching_feedback}), depending on the study, and specify the segment of the robot trajectory video to which the feedback applies. Participants can drag the video progress bar to the desired time point and set the start/end time by clicking the corresponding buttons. We ensured that the end time must be greater than or equal to the start time. The study's two teaching contexts are as follows:

\begin{figure}[b]
    \centering
    \includegraphics[width=0.48\textwidth]{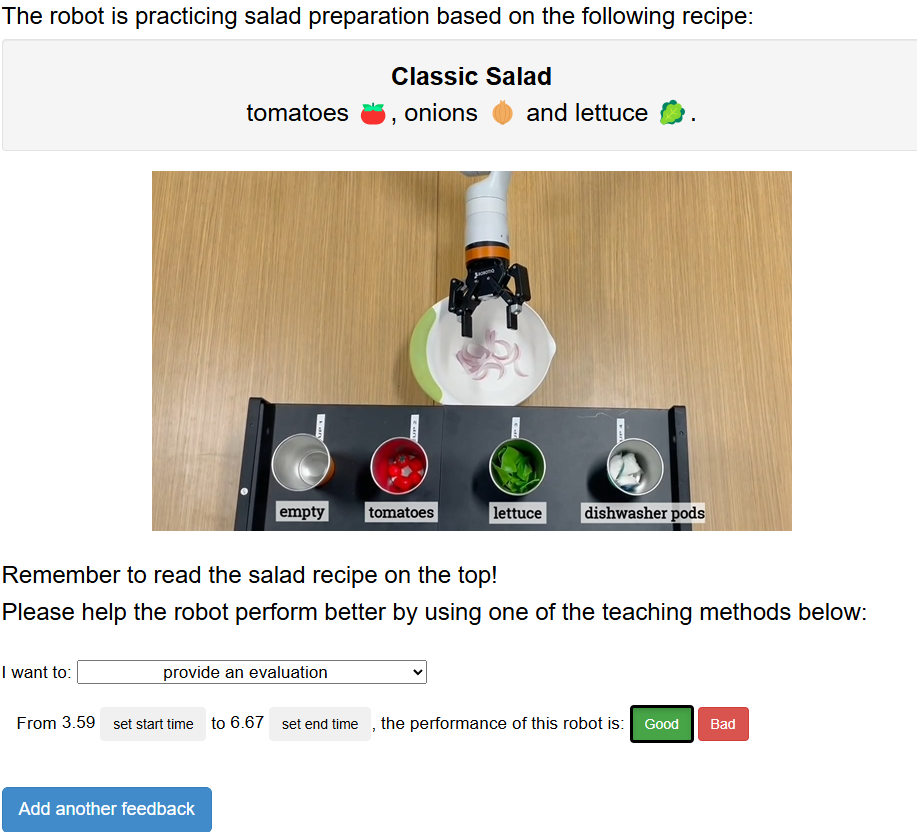}
        \caption{Interactive teaching interface. The participant watches a robot practicing video, specifies a feedback window and provides feedback to the robot learner.}
        \label{fig:teaching_interface}
\end{figure}

\textit{\textbf{Context 1: Forced-choice teaching}}: In this context, the participant needed to choose one feedback method from a set of common structured human feedback types used in interactive robot learning research. We determined our feedback set based on the taxonomy described in \cite{koppol2021interaction}. In this taxonomy, queries $q$, representing the data that prompts user responses, and feedback $f$, indicating the teaching signals provided by users, vary in size. To control for the effect of query size on human teaching, we excluded demonstrations, which have query size of zero ($len(q) = 0$), as well as preference and ranking teaching methods which have query sizes of $T \cdot 2$ and $T \cdot N$, respectively, where $T$ is the time horizon of a presented trajectory $\xi$, and $N$ is the number of trajectories in a query. 

Our final study interface incorporates 2 teaching methods: binary evaluation, which is $f_{eval}(\xi) = \{$\textit{good, bad}$\}$, and action instruction, which is $f_{instruct}(\xi) = \{$\textit{move forward, move backward, move to left, move to right, open gripper, close gripper, none}$\}$. We included a "none" option so that participants can choose it if they believe that the robot is performing well and still want to use instructions as their teaching method. For each piece of feedback, users also needed to define a feedback window specifying which part of the trajectory their feedback applies to.

\textit{\textbf{Context 2: Open-ended teaching}}: In this context, the participant was asked to provide open-ended feedback in natural language. They needed to define a feedback window and type their feedback in a text box. This allows people to express whatever they want to communicate to the robot learner.
\newline

For both contexts, a user had to provide at least one piece of feedback per trajectory. Similarly, for both contexts, users were able to add up to five pieces of feedback by clicking an "add feedback" button, as depicted in Fig. \ref{fig:teaching_interface}. This enables them to critique different parts of the trajectory or emphasize a certain feedback by repeating it. In context 1, the forced-choice teaching study, after the user gave their first piece of feedback, if they wanted to provide additional feedback, that feedback must also be via the same method. For instance, if a user chose to provide binary evaluative feedback for their first feedback, when they chose to add another piece of feedback, it would also be of the binary evaluative type. They could go back and change their minds at any point while critiquing that trajectory.

\subsection{Measures}
Here we define how we translated the three feedback dynamics, feedback granularity, feedback richness, and teaching time, into measures for analysis.

\subsubsection{Feedback granularity} To measure feedback granularity, we utilize the time ranges that participants provided to specify the segment of the robot trajectory their feedback applies to. Smaller time ranges indicate finer granularity of feedback, while larger time ranges denote less granularity. The finest granularity occurs when the start time equals the end time, and the largest when the time range spans the entire trajectory video.

\subsubsection{Feedback richness} 
\label{metric:richness}
In the forced-choice teaching context, feedback richness is quantified by the size of the response choice space, as described in \cite{koppol2021interaction}. For the evaluation method, where participants choose between "good" and "bad" buttons (binary feedback), the response choice space size is 2. For the instruction method, participants select an action from the robot's action set $A$, resulting in a response choice space size of $|A|$, in this case $|A|=7$. A larger response choice space size indicates greater feedback richness and vice versa.

In the open-ended teaching context, feedback richness is measured by calculating the word count of the text feedback. To control for an individual's overall verbosity, we normalize each participant's word count values by subtracting their minimum word count from the participant's own data. Longer text feedback corresponds to higher feedback richness, whereas shorter feedback implies lower richness. Naturally, there are cases where a participant's feedback may be long but contains information that could be said with fewer words. This does, however, serve as a proxy measure, providing an intuitive way for separate out broad feedback, such as "good job," and more specific instructional feedback about what the robot should do.

\subsubsection{Teaching time}
Teaching time refers to the duration of time spent by the participant in each trial. This is measured by recording the start and end times as the participant begins and finishes teaching the robot within a trial, then calculating the difference between these times.

\subsection{Experiment Procedure}
We conducted an online study and each experiment lasted $\sim$15 minutes. Each participant signed an informed consent form to confirm their eligibility (fluent English speaker, a United States resident, and at least 18 years old). The participant continued to complete a brief survey collecting their demographics, technology background, and previous robot experience. 

Next, the participant would use the interactive teaching interface described in Section \ref{sec:teaching_interface} to give feedback to the robot. The participant was randomly assigned to the forced-choice teaching context or the open-ended teaching context. In both cases, they would see six robot trajectories, as we described in Section \ref{sec:robot_errors}. To mitigate the ordering effect, we also randomized the order of robot trajectories using the Latin Square method \cite{kempthorne1955randomization}. For each trial, the participant watched a robot trajectory video and was prompted to provide feedback to improve the robot performance.

Finally, participants were asked to fill in a post-study questionnaire that included two Likert scale questions related to their teaching experience, and two open-ended questions asking about teaching strategies and general comments. The two Likert scale questions were: \textit{"I think the teaching interface is easy to use."} and \textit{"I can communicate my thoughts well using the teaching interface."} The two open-ended questions were: \textit{"What teaching strategy did you use for providing feedback to the robot?"} and \textit{"Do you have any overall comments about the study?"} Upon successful completion of all study components, the participants received a completion code for compensation.

\section{Results}
The procedure of our study was approved by the University Institutional Review Board (IRB), and we published our study on the Prolific platform \cite{palan2018prolific} to recruit participants. In both forced-choice and open-ended teaching contexts, participants were compensated \$4 for participating in the study. For the forced-choice teaching context, we recruited 32 participants (17 females, 15 males; aged 18-64). For the open-ended teaching context, we recruited 30 participants (24 females, 5 males, 1 non-binary; aged 18-70).

\textbf{Manipulation Check.}
To ensure that the design of the robot trajectories reflected different severity levels, we did a manipulation check on participants' subjective error severity ratings using Mann-Whitney U tests \cite{mcknight2010mann}. We observed that the severity ratings for successful trajectories ($M=1.556, SD=0.864$) were significantly lower than the ratings for erroneous trajectories ($M=4.012, SD=0.868$), \textit{$p$} < 0.001. We also conducted a Kruskal-Wallis test \cite{mckight2010kruskal} and validated that there exists 
some distinction in the severity ratings among all 4 erroneous trajectories ($H(3) = 20.984$, \textit{$p$} $<$ 0.001).\footnote{Details of error severity ratings of each robot trajectory can be found here: 

add dishpods (M=4.339), add nothing (M=4.048), miss tomatoes (M=3.903), drop onion (M=3.758), add lettuce correctly (M=1.839), add tomatoes correctly (M=1.274)}

\textbf{Quantitative Data Analysis Methods.} To analyze the data, we use both Bayesian statistics following the interpretation scheme presented in \cite{van_doorn_jasp_2021} and frequentist methods. All Bayesian tests were performed with a Cauchy prior distribution with $r=1/\sqrt{2}$. Tests for analyzing the effect of the presence of robot errors on human teaching (H1a, H2a, H3a) were done using Mann-Whitney U tests, since our data is not from a normal distribution as determined by a Shapiro-Wilk test. 
Tests for analyzing the effect of the severity of robot errors on human teaching (H1b, H2b, H3b) were done using Kendall rank correlation tests \cite{kendall1948rank}.

\subsection{Influence on Feedback Granularity} 
\subsubsection{\textbf{Results of forced-choice teaching}} 
\label{sec:nonverbal_granularity}
\begin{figure}[t]
    \centering
    \includegraphics[width=0.35\textwidth]{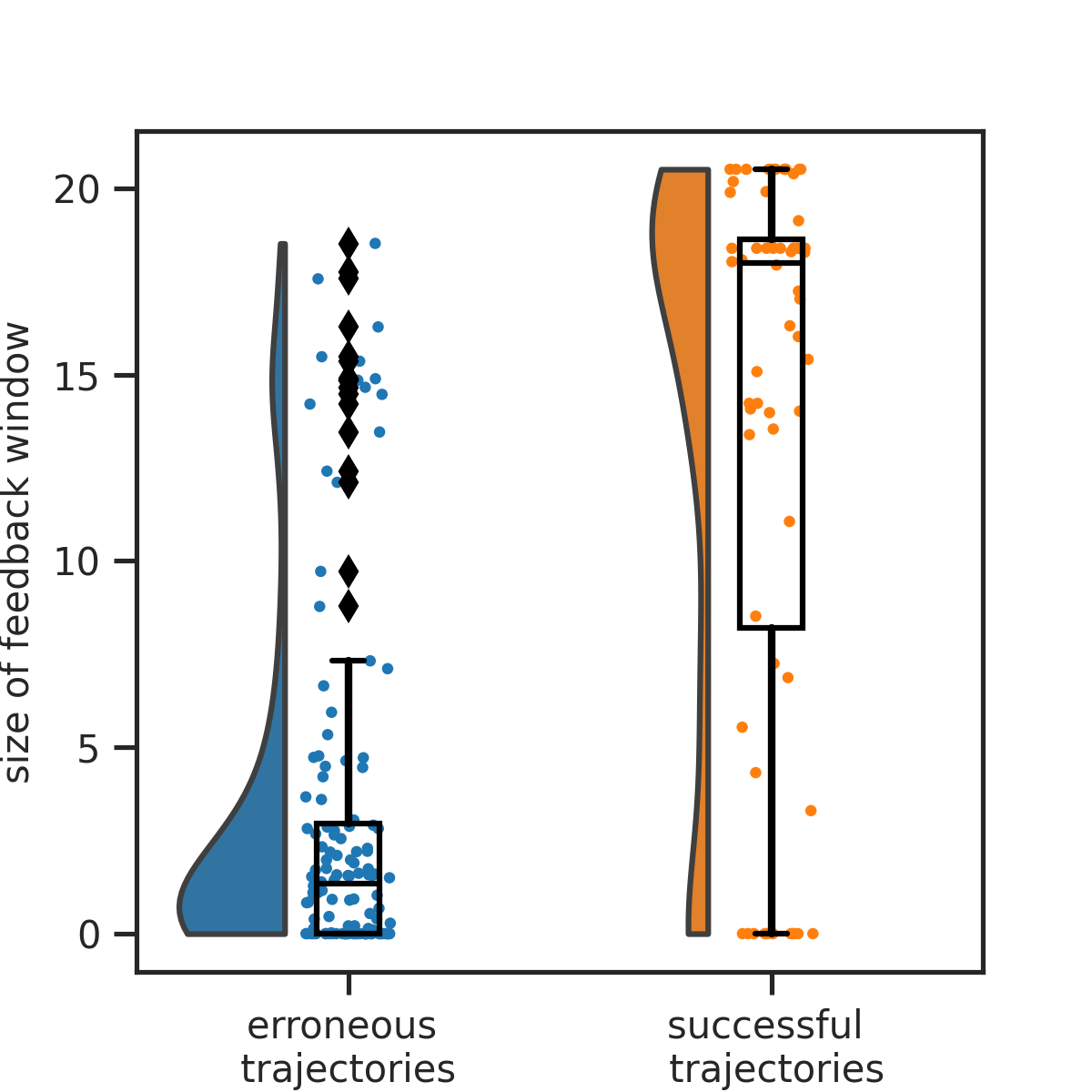}
        \caption{Feedback granularity data from forced-choice teaching, grouped by the presence of robot errors. }
        \label{fig:nonverbal_granularity}
\end{figure}

To test \textbf{H1a}, we categorized the feedback granularity data based on the presence of robot errors in each robot trajectory into two groups: successful trajectory group and erroneous trajectory group. Figure \ref{fig:nonverbal_granularity} visualizes the data. The results indicate that participants provided feedback with lower granularity for erroneous robot trajectories compared to successful ones (\textit{$p$} < 0.001, \textbf{$BF_{10}$} $>$ 10000), supporting \textbf{H1a}. Additionally, the severity of robot errors significantly correlates feedback granularity (\textit{$r_{\tau}$} $=$ -0.287, \textit{$p$} < 0.001, \textbf{$BF_{10}$} $>$ 10000), supporting \textbf{H1b}. Specifically, when the robot makes a more severe mistake, participants tend to specify a narrower credit assignment window to evaluate or instruct the robot.

\subsubsection{\textbf{Results of open-ended teaching}} 
\begin{figure}[t]
    \centering
    \includegraphics[width=0.35\textwidth]{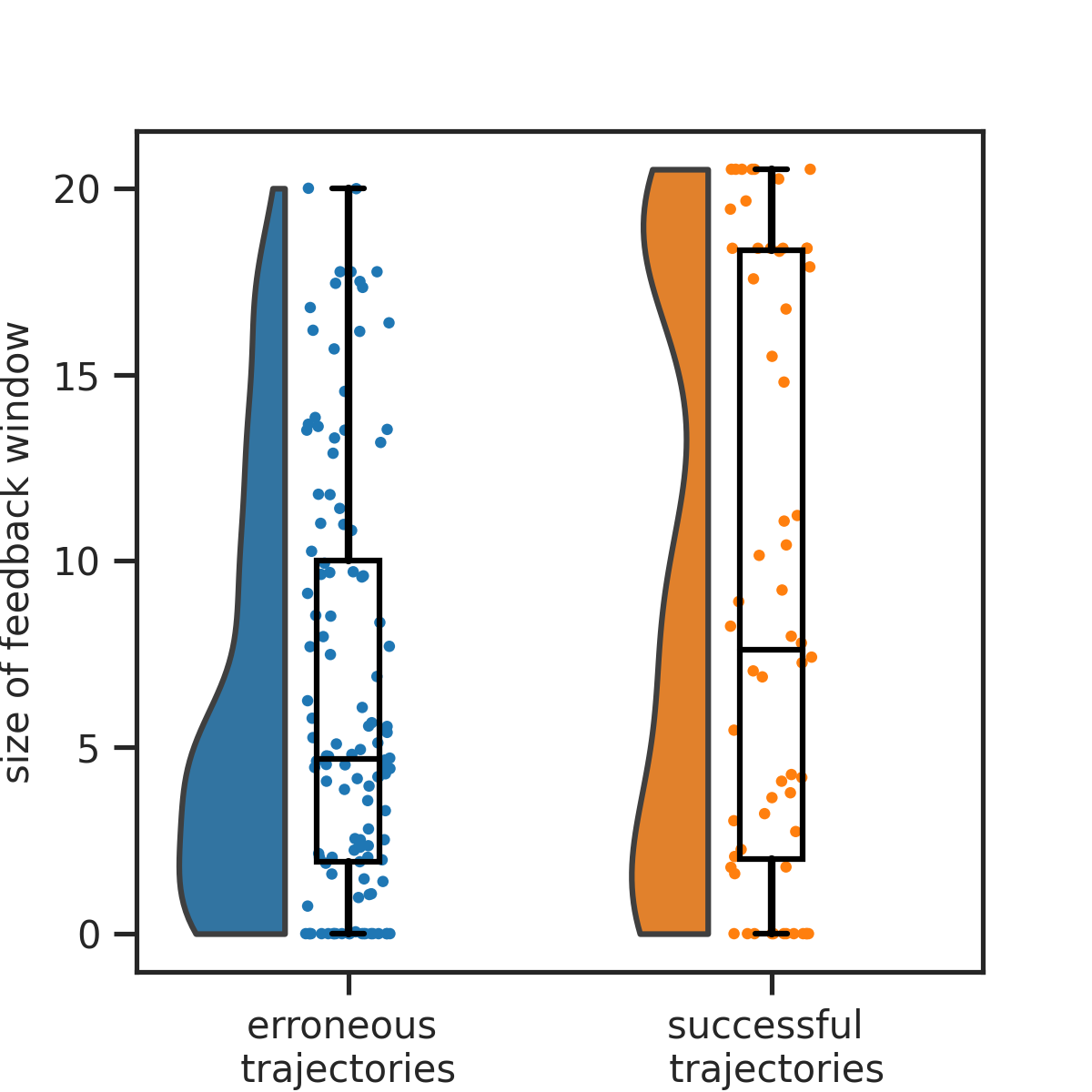}
        \caption{Feedback granularity data from open-ended teaching, grouped by the presence of robot errors. }
        \label{fig:verbal_granularity}
\end{figure}

Similar to the analysis in the section \ref{sec:nonverbal_granularity}, we found evidence that the presence of robot errors affects feedback granularity (\textit{$p$} = 0.043, \textbf{$BF_{10}$} = 2.336), which supports \textbf{H1a}. Fig. \ref{fig:verbal_granularity} illustrates that participants were inclined to specify a finer feedback window for erroneous trajectories, though the evidence is less strong compared to forced choice teaching data. We did not find a significant correlation between severity of errors and feedback granularity (\textit{$r_{\tau}$} = -0.089, \textit{$p$} = 0.111, \textbf{$BF_{10}$} = 0.468).

\subsection{Influence on Feedback Richness}

\subsubsection{\textbf{Results of forced-choice teaching}} 
\begin{figure}[b]
    \centering
    \includegraphics[width=0.35\textwidth]{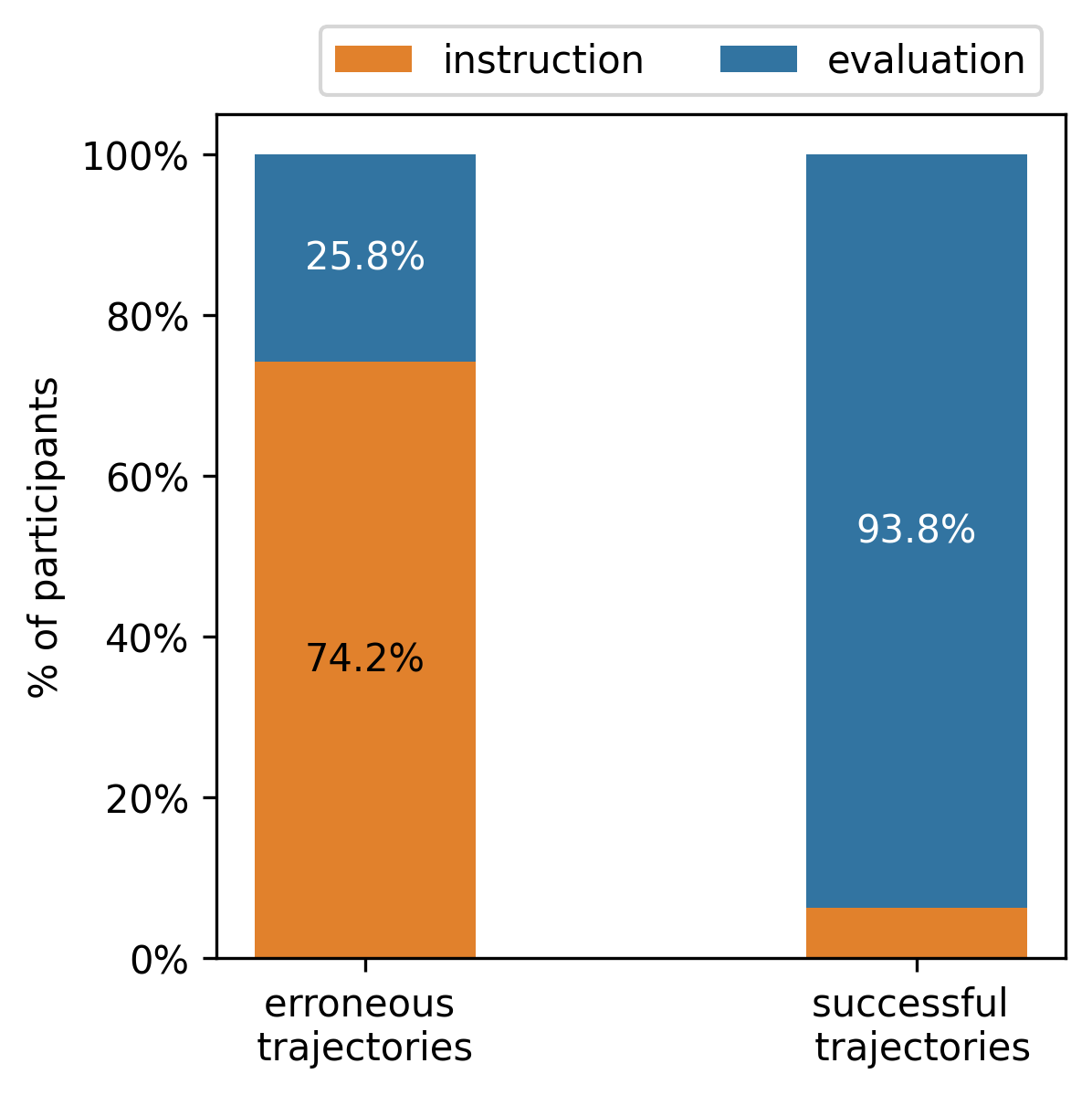}
        \caption{Feedback richness data from forced-choice teaching. In our study, instructive feedback is a richer form of feedback (larger size of response choice space) compared to evaluative feedback. Participants tended to evaluate successful trajectories and gave instructions for erroneous ones.}
        \label{fig:nonverbal_richness}
\end{figure}

In the forced-choice teaching context, we define feedback richness using the size of response choice space as described in section \ref{metric:richness}. 
We found that the presence of robot error significantly influences the feedback richness (\textit{$p$} < 0.001, \textbf{$BF_{10}$} $=$ 7857.122), which supports \textbf{H2a} (see Fig. \ref{fig:nonverbal_richness}). Additionally, we conducted a Mann-Whitney U t-test and found strong evidence that the severity of robot errors corresponds to the choices of teaching methods with different feedback richness. In particular, the severity ratings of the participants for the robot trajectories tended to be higher when they used the instructive teaching method compared to when they provided evaluative feedback (\textit{$p$} < 0.001, \textbf{$BF_{10}$} $>$ 10000). This evidence supports \textbf{H2b}.

\subsubsection{\textbf{Results of open-ended teaching}} 
\begin{figure}[b]
    \centering
    \includegraphics[width=0.35\textwidth]{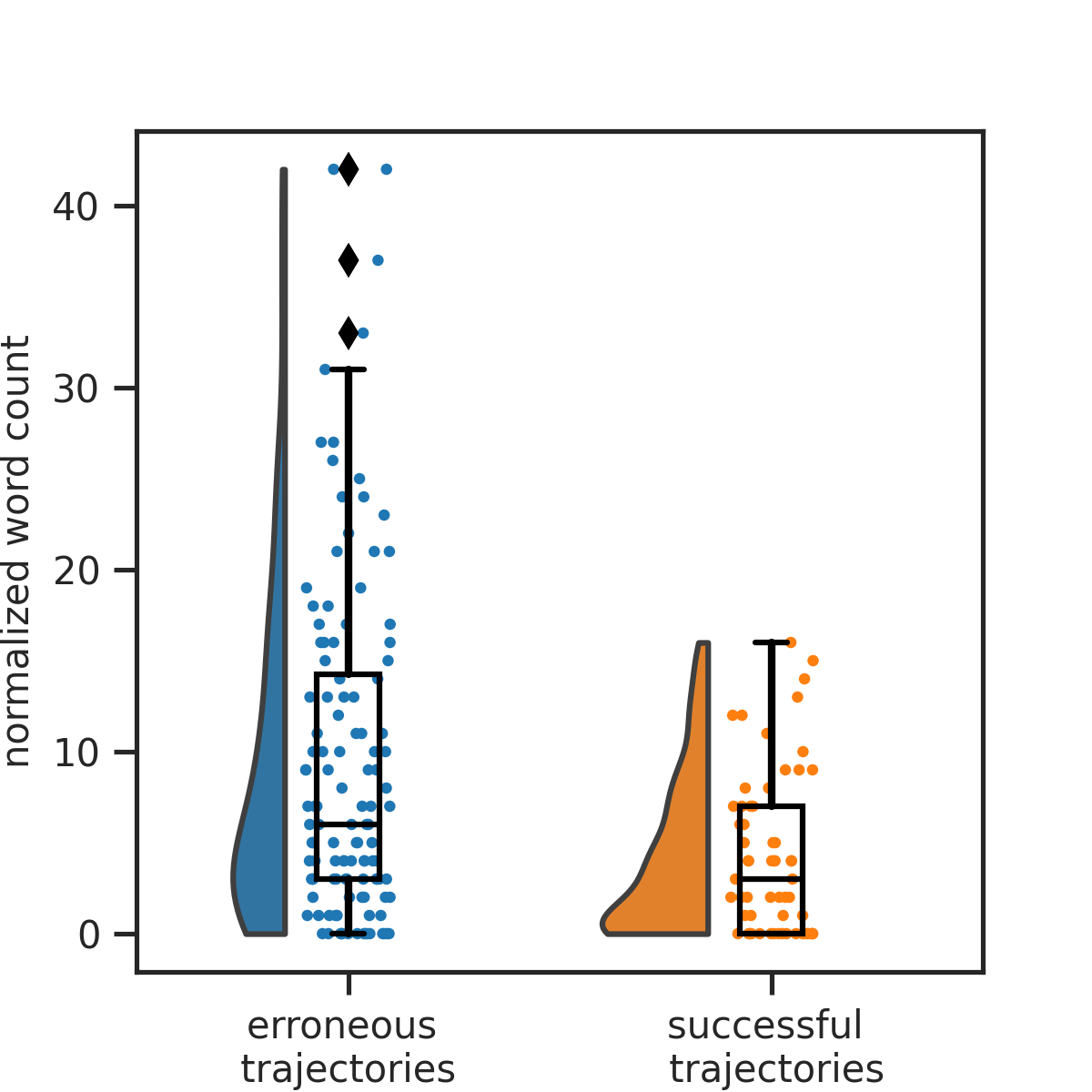}
        \caption{Normalized word count data from open-ended teaching, grouped by the presence of robot errors.}
        \label{fig:word_count_verbal}
\end{figure}

To measure the feedback richness, we calculated the word count for each language feedback as described in section \ref{metric:richness}. Figure \ref{fig:word_count_verbal} shows the distributions of word counts of language feedback used to teach the robot with both erroneous and successful trajectories. The results demonstrate that participants tend to provide longer language feedback for erroneous trajectories compared to successful ones (\textit{$p$} $<$ 0.001, \textbf{$BF_{10}$} = 63.238), which supports \textbf{H2a}. We also found anecdotal evidence \cite{beard2016using} for a correlation between error severity and feedback richness (\textit{$r_{\tau}$} = 0.119, $p = 0.037$, \textbf{$BF_{10}$} = 2.384), supporting \textbf{H2b}. 


\subsection{Influence on Teaching Time}

\subsubsection{\textbf{Results of forced-choice teaching}}
\begin{figure}[b]
    \centering
    \includegraphics[width=0.35\textwidth]{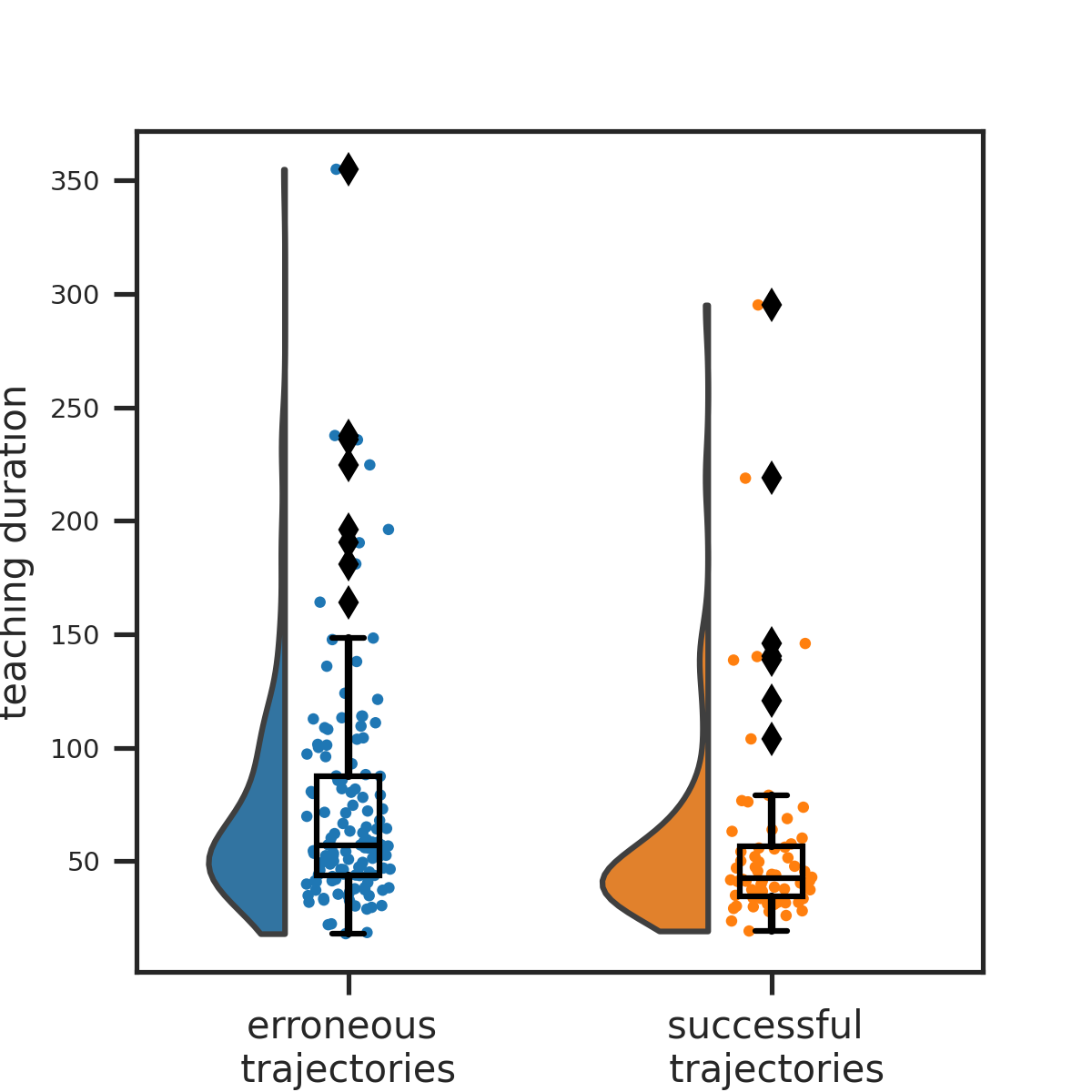}
        \caption{Teaching time collected in forced choice teaching, grouped by the presence of robot errors.}
        \label{fig:nonverbal_duration}
\end{figure}

We found strong evidence that the presence of robot errors has an impact on teaching time (\textit{$p$} $<$ 0.001, \textbf{$BF_{10}$} $=$ 11.729), which supports \textbf{H3a} (see Fig. \ref{fig:nonverbal_duration}). It reveals that in forced choice teaching, participants spent more time to teach erroneous trajectories compared to successful ones.
Moreover, the severity of robot errors correlates positively to teaching time 
(\textit{$r_{\tau}$} = 0.141, \textit{$p$} $=$ 0.009, \textbf{$BF_{10}$} = 6.103), supporting \textbf{H3b} as well. That is to say, participants spent more time teaching the robot when it made a more severe error. 

\subsubsection{\textbf{Results of open-ended teaching}}

We failed to find a significant result that the presence of robot errors influences the amount of time participants spent on teaching in open-ended teaching (\textit{$p$} $=$ 0.672, \textbf{$BF_{10}$} $=$ 0.203). Similarly, we did not find any evidence that the severity of robot correlates to teaching time (\textit{$r_{\tau}$} = -0.030, \textit{$p$} $=$ 0.587, \textbf{$BF_{10}$} = 0.116).

\subsection{Overall Teaching Experience in Forced-choice and Open-ended Teaching}
In the post-study questionnaire, each participant was asked to rate the usability and expressiveness of the teaching interface using a 5-point likert scale. To understand whether there is any differences between the data collected from forced-choice and open-ended teaching, we conducted two-sample t-tests. We failed to find significant differences in interface usability  (\textit{$p$} = 0.079, \textbf{$BF_{10}$} = 0.983). However, there was a significant difference in interface expressiveness (\textit{$p$} = 0.011, \textbf{$BF_{10}$} = 4.334). This finding indicates that participants perceived language feedback to be more expressive than structured evaluative and instructive feedback. Given that usability was not perceived differently by participants, a promising research direction could be to develop more expressive teaching interfaces for forced choice teaching scenarios.

\section{Discussion}

In this paper, we demonstrate that the presence and severity of robot errors affect three dimensions of human teaching dynamics: feedback granularity, feedback richness, and teaching time. Compared to a successful example, when the robot displays erroneous behavior, people are inclined to use a narrower feedback window, provide more detailed information, and spend more time teaching the robot. In general, this tendency becomes more pronounced as the severity of the robot error increases.

Interestingly, the effects of robot errors on human teaching are less pronounced in the open-ended teaching context than in the forced choice one. For instance, in the open-ended teaching, we saw a bimodal distribution of the feedback granularity data from the successful trajectory group (see Fig. \ref{fig:verbal_granularity}). Some participants provided feedback over a long trajectory, while others focused on a small segment of the robot's actions or even a single action. This variability may be attributed to the nature of language feedback, which is less structured than binary feedback or action instructions, and to the diverse metrics that people use to assess robot performance. Although the robot may perform well in terms of task completion, participants might identify sub-optimalities in other areas, such as task planning (\textit{P1-open-ended: "The robot should have started off with tomatoes, but this will still work."}) and fine manipulations (\textit{P4-open-ended: "When pouring ingredients into the bowl, you want to make sure that you are pouring directly into the center of the bowl."}).

\begin{figure}[t]
    \centering
    \includegraphics[width=0.52\textwidth]{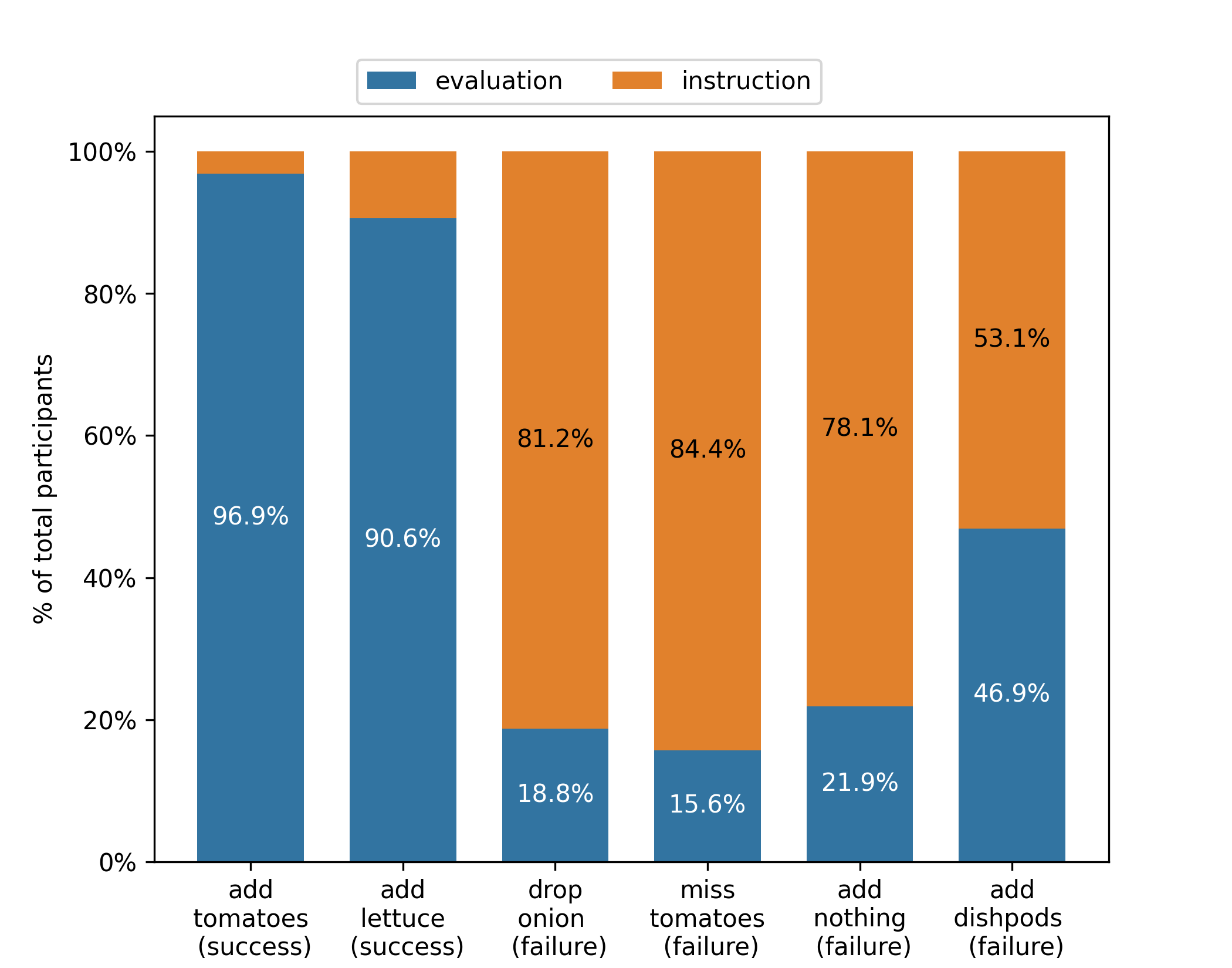}
        \caption{Choices of teaching methods in forced choice teaching context, grouped by robot trajectories. }
        \label{fig:res_teaching_modes}
\end{figure}

Language feedback allows flexibility in expressing teaching information but may be less straightforward for robot learners to interpret. From the robot learning perspective, ideal human feedback should specify what the error is, when it occurs, how it happens, and potential solutions to fix it. However, many participants only provided information related to error detection, such as \textit{"P6-open-ended: the robot missed the bowl, spilling onions"} and \textit{"P4-open-ended: Dishwasher pods do not belong in this salad"}. This tendency introduces ambiguity to robot learners as they may not be able to reason and extract other key information required for improving the learning performance. Future work could focus on detecting and translating ambiguity in language feedback, for instance, designing a robot querying mechanism that asks for additional error explanation and recovery information to better facilitate robot learning.

Additionally, the nature of robot errors plays a role on human teaching dynamics. As illustrated in Fig. \ref{fig:res_teaching_modes}, when the robot made a huge mistake, such as adding dishwasher pods to the salad bowl, binary evaluative feedback was used more often among the participants compared to the other erroneous trajectories. This may be attributed to the way people perceive errors - not all errors are perceived equally. While some participants used action-based instructions to correct the dishwasher pod error by directing the robot to grab the cup with the correct ingredients, others viewed it as a complicated conceptual mistake. They believed that the robot did not understand dishwasher pods are inedible and thus chose to use evaluative "bad" feedback signals to communicate this knowledge to the robot (\textit{"P12-forced-choice: I liked to give clear instructions for when/how the robot hand should move. But in the dishwasher pod case, it is not about movements, more like not knowing a concept. It's hard to teach using a series of instructions, so I gave bad feedback"}).
This implies that choices over the type of feedback can contain information related to teaching strategies, which may lead to differently-aligned learning systems \cite{conitzer2024social}. Future work should further examine behavioral effects of different feedback types.

Our data also suggest that people may employ teaching strategies found in human-human teaching. In the open-ended teaching context, some participants adopted a positive reinforcement strategy, which is widely used in educational fields \cite{maag2001rewarded, arista2018types}. For example, even though the robot failed to grab the cup with tomatoes, a few participants still offered encouragement, saying, \textit{"P8-open-ended: you'll get it next time."} and \textit{"P14-open-ended: You were in the right spot, but you have to clamp onto the cup. All your motions after leaving the cup of tomatoes were great."} This use of positive reinforcement highlights the natural tendency to motivate and support the learner, even in the context of robot training. When designing HITL learning algorithms, robots should be aware of this natural tendency, both to accurately interpret encouraging feedback and to respond to human teachers in a human-like way.

\textbf{Limitations and Future Work.} 
There is a wide range of teaching contexts and scenarios where robot error could impact a user's teaching style and behavior. In this work, we chose to focus on an online crowdsourcing setting where users provide feedback to prerecorded robot trajectories. This setting is particularly important as large robot-behavior models begin to develop and incorporate RLHF into their training. However, the results and takeaways in this context may not apply to in-person human-robot teaching scenarios where the interaction may span a longer period and where the robot may be adapting its behavior in real-time. In that case, a user's perception of the robot error may change as a function of how the robot improves at the task for instance. This setting warrants its own investigation as well as analysis as to how it differs from the online setting. 

Future work should investigate potential causes other than robot errors that influence teaching dynamics; for example, the nature of the task the robot is performing or legibility of the robot's motions. Future work should also study how a user's choice of teaching method is affected by robot error when a broader range of feedback modalities are available. In our forced-choice teaching study, for instance, we provided users with a choice of only binary evaluation or single-action instructions. This choice allowed us to compare the two feedback methods of evaluation and instruction with distinct levels of feedback richness and specificity. However, there are other ways of collecting feedback, such as collecting full-trajectory demonstrations, querying for preferences, or learning from implicit feedback like facial expressions, which can further enhance teachers' expression and choices. Lastly, a limitation with regard to our evaluation is that our analysis of feedback richness using word-count did not capture all of the nuances of the language feedback that was provided. For example, an equally long sentence could discuss multiple errors or focus on one error. This difficulty also prevented us, at least in this work, from more directly comparing the feedback results from forced choice and open-ended teaching. Defining feedback richness in a measurable way that captures both the user's intent as well as the amount of relevant information to the learning process itself is an on-going area of research. 

\vspace{-6pt}
\section{Conclusion} 
As researchers design human-in-the-loop learning algorithms and as these systems are deployed, it is crucial to ensure that the teaching process is efficient, accessible, and comfortable. Robot errors are and will continue to be an inevitable part of this process. Developing a greater understanding of how these errors influence the teaching dynamics enables us to create better teaching experiences as well as learning algorithms. To that end, this paper presented results about how error can affect three dimensions of human teaching dynamics: feedback granularity, feedback richness, and teaching time. We studied this through the lens of an online teaching scenario and found that errors not only influence these teaching dynamics but often do so in a predictable way. Although there is more to be explored regarding how robot errors impact teachers in other human-robot teaching settings, this paper lays the groundwork for studying these phenomena and provides insights that can be applied to the design of human-in-the-loop learning systems.

\begin{acks}
This work was supported in part by the US National Science Foundation (IIS-2132887). We thank Ziqi Chen for designing the study interface and the anonymous reviewers for their valuable feedback. 
\end{acks}

\bibliographystyle{ACM-Reference-Format}
\bibliography{ref}

\end{document}